\documentclass[manuscript]{acmart}

\makeatletter                   
\def\mdseries@tt{m}             
\makeatother                    

\usepackage[draft=true]{minted} 
\usepackage{color}
\usepackage{hyperref}           
\hypersetup{
    colorlinks=true,
    linkcolor=blue,
    filecolor=red,      
    urlcolor=magenta,
    breaklinks=true,            
}
\usepackage{breakurl}           

\AtBeginDocument{%
  }
\setcopyright{acmcopyright}
\copyrightyear{2022}
\acmYear{2022}
\acmDOI{XXXXXXX.XXXXXXX}

\acmConference[RecSys'22 CONSEQUENCES Workshop]{Recommender Systems Conference, CONSEQUENCES Workshop}{2022}{Seattle, WA}
\acmPrice{15.00}
\acmISBN{978-1-4503-XXXX-X/18/06}

\usepackage{enumitem}
\usepackage[utf8]{inputenc}
\usepackage{graphicx}
\graphicspath{ {plots/} }
\usepackage{array}

\usepackage[utf8]{inputenc} 
\usepackage[T1]{fontenc}    
\usepackage{booktabs}       
\usepackage{amsfonts}       
\usepackage{nicefrac}       
\usepackage{microtype}      
\usepackage{amsmath,amsfonts}
\usepackage{algorithm}
\usepackage[noend]{algpseudocode} 
\usepackage{graphicx}
\usepackage{color, colortbl}
\usepackage{subfig}
\usepackage{bbm}
\usepackage{soul}
\usepackage{xcolor}

\definecolor{LightGreen}{rgb}{0,1,0}
\definecolor{Green}{rgb}{0,0.5,0}
\graphicspath{{figures/}}

\newcommand{\houssam}[1]{\textcolor{green}{Houssam: #1}}

\renewcommand{\P}{\mathbb{P}}

\newcommand{\mc}{\mathcal}

\begin{document}
\sloppy
\title{Adaptive Experimental Design and Counterfactual Inference}

\author{Tanner Fiez}
\author{Sergio Gamez}
\author{Arick Chen}
\author{Houssam Nassif}
\author{Lalit Jain}
\additionalaffiliation{University of Washington}
\affiliation{\institution{Amazon}}


\begin{abstract}
Adaptive experimental design methods are increasingly being used in industry as a tool to boost testing throughput or reduce experimentation cost relative to traditional A/B/N testing methods. 
This paper shares lessons learned regarding the challenges and pitfalls of naively using adaptive experimentation systems in industrial settings where non-stationarity is prevalent, while also providing perspectives on the proper objectives and system specifications in these settings. 
We developed an adaptive experimental design framework for counterfactual inference based on these experiences, and tested it in a commercial environment.
\end{abstract}


\maketitle

\section{Introduction}
\label{sec:intro}
A/B/N testing is a classic and ubiquitous form of experimentation that has a proven track record of driving key performance indicators within industry~\citep{kohavi2020trustworthy}.
Yet, experimenters are steadily shifting toward \textit{Adaptive Experimental Design} (AED) methods 
with the goal of increasing testing throughput or reducing the cost of experimentation.
AED promises to use a fraction of the impressions that  traditional A/B/N tests require to yield high confidence inferences or to directly drive business impact. 
In this paper, we share lessons learned regarding the challenges and pitfalls of naively using adaptive experimentation systems in industrial settings where non-stationarity is the norm rather than the exception. Moreover, we provide perspectives on the proper objectives and system specifications in these settings.
This culminates in a high level presentation of an AED framework for counterfactual inference. To provide a robust and flexible tool for experimenters with performance certificates at minimal cost, our methodology combines cumulative gain estimators, always-valid confidence intervals, and an elimination algorithm.

\section{A Case Study}
\label{sec:case_study}
\begin{figure}[t!]
    \centering
    \subfloat{\includegraphics[width=.5\textwidth]{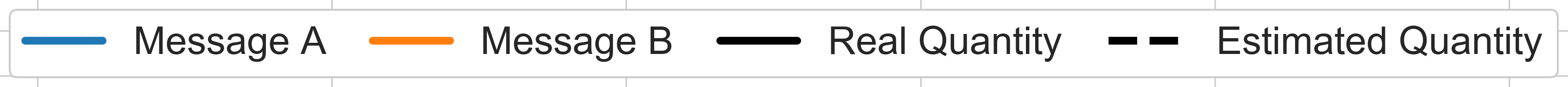}}
    \setcounter{subfigure}{0}
    
    \subfloat[Daily play probability]{\includegraphics[width=.25\textwidth]{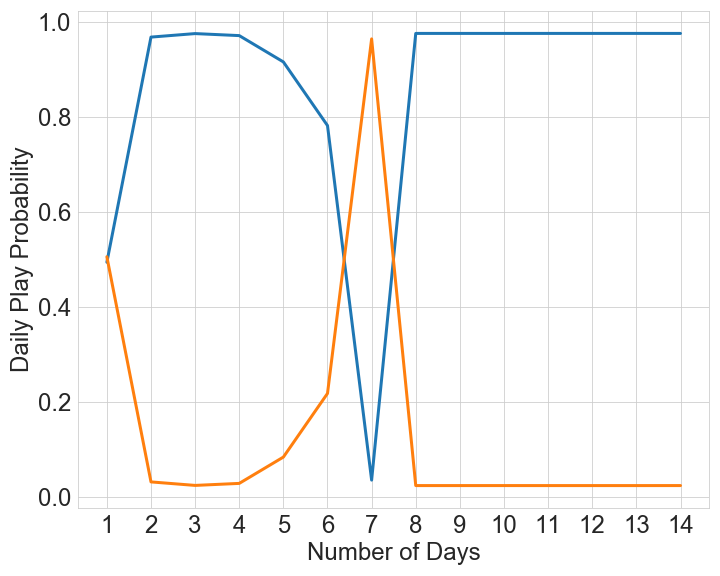}}
    \subfloat[Daily mean]{\includegraphics[width=.25\textwidth]{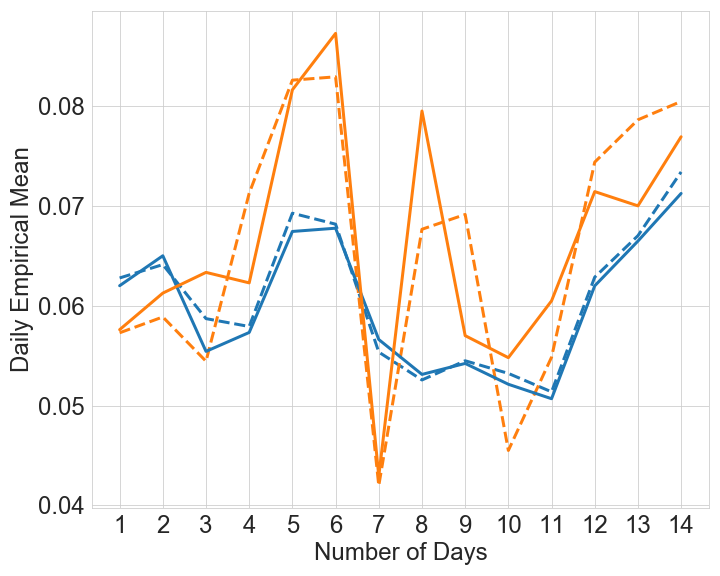}}
    \subfloat[Running empirical mean]{\includegraphics[width=.25\textwidth]{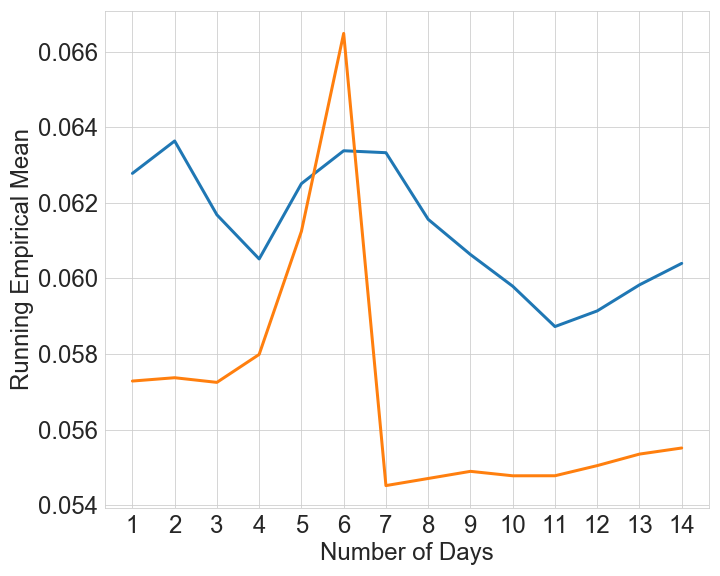}\label{fig:figc}}
    \subfloat[Cumulative gain]{\includegraphics[width=.25\textwidth]{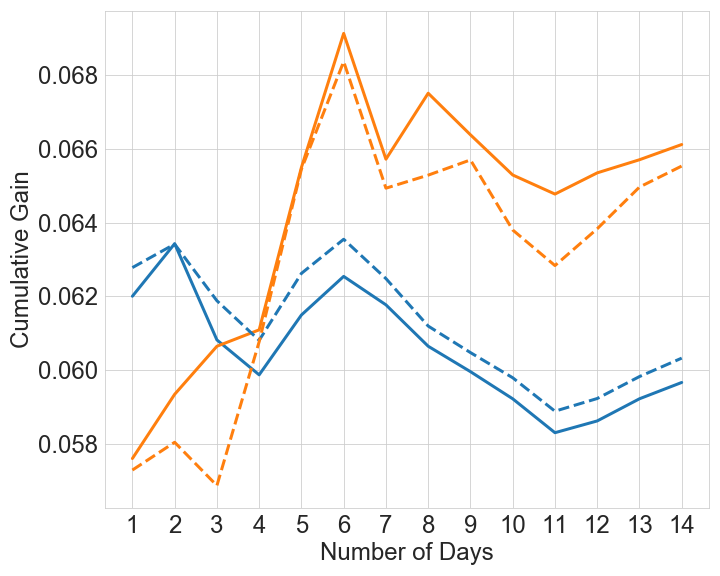}\label{fig:figd}}
    
    \caption{Case study of time-variation and adaptive allocations causing Simpson's paradox.}
     \label{fig:ex}
\end{figure}

Imagine a setting where on a retailer web page, a marketer has been running a message $A$ for the last year and now wants to test whether message $B$ beats $A$. At the start of the experiment the messages are initialized with a default prior distribution, and then at each round a Thompson sampling bandit dynamically allocates traffic to each treatment, playing each message according to the posterior probability of its mean being the highest~\cite{SawantHVAE18}. After day 8, the algorithm directs most traffic to message $A$  (see Figure~\ref{fig:ex}). On day 14, the experimenter needs to decide whether $A$ has actually beaten $B$. They conduct a paired t-test which, somewhat surprisingly, does not produce a significant $p$-value. As the bandit shifted all traffic to message $A$, not enough traffic was directed to message $B$, diminishing the power of the test. The experimenter is forced to conclude that they can not reject the null hypothesis that there is no difference between the messages. A few days later, the experimenter, who is still perplexed, looks at the daily means and is then shocked to see that on most days, $B$ tends to have a higher empirical mean than $A$, which disagrees with the bandit's beliefs.

To understand this behavior, note that in Figure~\ref{fig:figc} the cumulative success rate of $A$ is exceeding that of $B$, leading the algorithm to put all its traffic on $A$. This phenomenon where the total success rate shows a different direction than daily comparisons is referred to as \textit{Simpson's Paradox}, and occurs in settings where the traffic is dynamically allocated to arms whose means change over time~\citep{kohavi2020trustworthy}. At an intuitive level, the experimenter has perhaps made a Type I error by trusting the algorithm and choosing arm $A$. Indeed, during the time period from day 8 to day 14, the algorithm decided to put more traffic on arm $A$, exacerbating Simpson's paradox. Convinced by it's own bad decision, the algorithm then chooses a bad traffic allocation which further exacerbates the problem and leads to a vicious cycle.

\section{Lessons learned} 
\label{sec:learning}
The case study from Section~\ref{sec:case_study}, though simple, demonstrates many of the challenges and pitfalls of naively using adaptive experimentation systems in industrial settings where time variation is the norm rather than the exception. We now dive deeper into some of these concerns and also provide thoughts on objectives and specifications in these settings.

\noindent\textbf{Regret Minimization Isn't Enough.} The fundamental goal of experimentation is to test hypothesis and deliver results that allow for \textit{future iterations}~\cite{biswas2019seeker}.  As a result, it is important that experimentation procedures give the experimenter the ability to arrive at valid and measurable inferences. In settings where the experimenter wants to learn the best treatment, optimal regret minimization procedures (in stochastic settings) tend to have much lower power (as they are far from a balanced allocation) and can take a significantly longer time to return the identity of the best arm with high probability~\citep{audibert2010best, degenne2019bridging}. 
In addition regret minimization procedures lead to biased estimates of empirical means \citep{shin2019bias}. 

\noindent\textbf{Be Wary of the Batch.} Most experimental systems use batched model updates (daily or weekly). In the example from Section~\ref{sec:case_study}, the traffic was not constant daily (not shown), so an update on one day can have an undue impact on the rest of the experiment time. In experiments over short horizons, this implies that observations on the first few days can have a disproportionate impact on the traffic allocation and also the subsequent inferences that are made.

\noindent\textbf{Identify the Counterfactual Best.} 
Naively assuming stationarity and using empirical mean estimates can lead to faulty inferences. In general, regret minimization is a well defined objective, however it may fail to deliver the inference that the marketer is intuitively looking for, that is, the identity of the best-arm. Furthermore, in settings where arm means are shifting over time, it is challenging to define this notion as the mean performance of an arm and the identity of the best arm may change daily. 
To bridge this gap, our proposed objective is to \textit{identify with high probability the treatment that would have obtained the highest possible reward, if all traffic had been diverted to it}. This counterfactual metric is known as the \textit{cumulative gain}. 
Figure~\ref{fig:figd} demonstrates the cumulative gain over time for the case study. With the exception of~\citep{abbasi2018best}, we believe that this objective has hardly been considered in the best-arm identification literature. 

\noindent\textbf{Always Valid Inference.}  In traditional A/B testing, the experiment horizon is fixed ahead of time (generally based on a minimum detectable effect size the experimenter is interested in detecting) and then a test for significance is used at the end of the experiment. Monitoring, or stopping the experiment, based on $p$-values computed during the experiment (a process known as \emph{p-hacking}) is heavily frowned upon as it leads to Type 1 error inflation~\citep{johari2017peeking}. Recent work in the experimental space that builds upon ideas of~\citet{robbins1970statistical}, has lead to generalizations of the $p$-value known as \emph{always-valid p-values} which are slightly inflated and can safely be sequentially monitored~\cite{jamieson2014lil, johari2017peeking, howard2021time}. This capability is critically important in practice to allow for early stopping and to ensure valid inferences are drawn. 

\noindent\textbf{The Best of Three Worlds.} Though optimal regret minimization procedures fail to provide valid inferences and tend to identify the best arm more slowly, we still would like to minimize the opportunity cost of experimentation. Thus experimentation systems should try to provide the best of three worlds: identification of the counterfactual best, mitigation of opportunity cost, robustness to arbitrary time variation. In completely adversarial settings we can't hope to have all three~\citep{abbasi2018best}, but real life settings mostly live somewhere between fully stochastic and fully adversarial.

\section{Robust and Adaptive Counterfactual Inference}
\label{sec:anduril}
We now present an AED methodology for counterfactual inference that is simple, mitigates many of the concerns raised regarding non-stationarity, and adheres to the preferred objectives and system specifications.

\noindent\textbf{Experimentation Setting.}
\label{sec:setting}
We consider an experimentation setting over $k\geq 2$ arms with potentially non-stationary daily success rates and batch feedback. On any day $t\geq 1$, each customer that arrives is shown an arm $i\in [k]$ with probability $p_{i,t}$, where the sampling distribution $p_t = (p_{1,t}, \cdots, p_{k,t})\in \Delta^{k}=\{p_t\in \mathbb{R}^k : p_{i, t}\geq 0 \ \forall i\in [k], \sum_{i=1}^k p_{i, t}=1\}$ is chosen by an algorithm dependent on the observations prior to day $t$. Denote by $n_{i, t}$ the number of impressions for arm $i\in [k]$ on day $t$ and the total day traffic by $n_t= \sum_{i\in [k]}n_{i, t}$. 
Let $r_{i, t}$ and $\widehat{\mu}_{i, t} :=r_{i, t}/n_{i, t}$ denote the total reward and empirical mean on day $t$ for any arm $i\in [k]$, respectively. We further assume that the mean of an arm $i\in [k]$ on any day $t\geq 1$ is fixed over the day and denote it by $\mu_{i, t}\in [0,1]$.  Thus, conditional on the allocation $n_{i,t}$, $r_{i,t}\sim \text{B}(n_{i,t}, \mu_{i,t})$. 

\noindent\textbf{Estimation: Cumulative Gain.}
\label{sec:est}
As discussed previously, assessing the performance of arms using empirical mean estimates can lead to faulty inferences in the presence of dynamic traffic allocations and non-stationarity due to biases and the ill-defined nature of the underlying quantity being estimated. 
To overcome this challenge, we consider the counterfactual metric known as the cumulative gain that measures the performance of an arm had it received all of the impressions.
Formally, for any arm $i\in [k]$, the \textit{cumulative gain} after $t\geq 1$ days is $G_{i, t} := \sum_{s=1}^t n_s \mu_{i,s}$.
We can build an estimator for this metric using inverse probability weighting. 
Indeed, an unbiased cumulative gain estimator is $ \widehat{G}_{i,t} = \sum_{s=1}^t r_{i,s}/p_{i,s}$.
Unlike an empirical mean, the cumulative gain estimator will never suffer from \textit{Simpson's paradox}.

\noindent\textbf{Algorithm: Elimination on Cumulative Gain via Sequential Monitoring.}
\label{sec:elim}
Given the cumulative gain metric, our objective is to identify the arm with the maximum cumulative gain with high probability while also minimizing regret. Motivated by the experimentation setting, metric and objective, we have arrived at the elimination style algorithm presented in Algorithm~\ref{alg:ae} as a procedure for AED and counterfactual inference.
As input, Algorithm~\ref{alg:ae} takes a set of arms $[k]$, a confidence parameter $\delta$ (normally set to 0.1), and a ``settling period'' $\tau$.  On each day, an active set of arms $\mc{A}$ is maintained and each is shown with equal probability $1/|\mc{A}|$. At the end of each day, after the settling period, it removes any arms that it can verify are statistically worse than an existing arm based on the cumulative gain metric. 
Specifically, Algorithm~\ref{alg:ae} eliminates using an always-valid confidence interval \cite{johari2017peeking, howard2021time, jamieson2018bandit}. That is, for each pair of arms $i,j\in [k]$ an always-valid confidence interval guarantees that 
\[\P(\exists t\geq 1, i,j\in [k]: |(\widehat{G}_{i,t} - \widehat{G}_{j,t}) - (G_{i,t} - G_{j,t})| \geq \phi(i,j,t,\delta)) \leq \delta. \]
 If each $n_t$ is sufficiently large and each arm receives enough traffic, we can invoke the CLT and as an approximation employ the mixture sequential probability ratio test (MSPRT)~\cite{johari2017peeking, robbins1970statistical} to define the always-valid confidence interval
 \[\textstyle \phi(i,j,t,\delta) := \sqrt{(V_t(i, j)+\rho)\log\big((V_t(i, j)+\rho)/(\rho\delta^2)\big)} \quad \text{where}\quad V_t(i, j)=\sum_{s=1}^t n_s\big(\widehat{\mu}_{i,s}(1-\widehat{\mu}_{i,s})/p_{i,s}+\widehat{\mu}_{j,s}(1-\widehat{\mu}_{j,s})/p_{j,s}\big)\] and $\rho >0$ is a fixed constant. 
Motivated by the existence of the always valid confidence interval, Algorithm~\ref{alg:ae} eliminates an arm $j\in [k]$ on some day $t\geq 1$ when there exists an arm $i\in [k]$ such that $\widehat{G}_{i,t}-\widehat{G}_{j,t} -\phi(i,j,t,\delta) > 0$.

\begin{algorithm}[t]
\footnotesize
\begin{algorithmic}[1]
\State{\textbf{Input} Arm set $[k]$, Confidence $\delta\in (0, 1)$, Settling Time $\tau\geq 1$}
\State{\textbf{Initialize}  $\mc{A} \gets [k], t\gets 1$}

\While{$|\mc{A}| > 1$} 
\State{On day $t$, set $p_{i,t} = 1/|\mc{A}|$ for all $i\in \mc{A}$ and for each customer $s\leq n_t$, show arm $i$ with probability $p_{i,t}$}
\If{$t\geq \tau$}
\State{$\mc{A}\gets \mc{A}\setminus\{j\in \mc{A}:\exists i\in \mc{A}, \widehat{G}_{i,t} - \widehat{G}_{j,t} - \phi(i,j,t,\delta/k)>0\}$}

\EndIf
\State{$t\gets t+1$}
\EndWhile
\State{Return $\mc{A}$}
\end{algorithmic}
\caption{Successive Elimination on Cumulative Gain}
\label{alg:ae}
\end{algorithm}

\noindent\textbf{AED System Guarantees.} 
In the stochastic stationary setting or the constant gap setting, Algorithm~\ref{alg:ae} reduces to a version of the successive elimination algorithm~\cite{even2006action}.  In this case we have a guarantee that the best arm will be returned with probability greater than $1-\delta$ in a number of samples not exceeding $O(\log(k/\delta)\sum_{i=1}^k \Delta_i^{-2})$ and with an instance-dependent regret of no more than $O(\log(k/\delta)\sum_{i=1}^k \Delta_i^{-1})$, both of which are near-optimal~\cite{kaufmann2016complexity, jamieson2014lil, zhaoqi2022InstanceOptimal}. 
In the general non-stationary setting it is more difficult to make a strong statement about Algorithm~\ref{alg:ae}'s performance, beyond that if an arm is eliminated then there exists an arm with a higher cumulative gain in the active set at that day. Note that the downside of elimination in a time-varying setting is that an arm that is eliminated because it is sub-optimal today, could potentially be the best performing arm in the future.
However, from a practical perspective, eliminating arms once they become sub-optimal is an easily interpretable solution that allows practitioners to hone in on a winner. 
In addition, as the number of arms shrink, the remaining arms acquire more samples, increasing power over time.

\section{Discussion \& Future Directions}
In this work, we discuss the challenges of applying AED methods in practice, provide thoughts on objectives and specifications of such systems, and present the approach we have arrived at using AED for counterfactual inference. Our methodology is perhaps most closely linked to the best-of-both-worlds setting and the P1 algorithm~\cite{abbasi2018best}, where the experimenter in each round plays each arm with some positive probability and declares a winner when there is one arm whose lower confidence bound is greater than the upper confidence bound of each other arm.
We take a far more aggressive approach that effectively guarantees that in the stochastic setting we will recover the best possible sample complexity, but give up on strong guarantees in the adversarial setting. Based on our experiences, this appears to be reasonable. This work brings up a number of interesting future work directions including exploring combinatorial settings for multivariate testing, and experimentation dynamics leveraging priors on the probability of launching successful treatments to more effectively balance identification and regret objectives~\cite{nabi2022EB}.

\bibliographystyle{plainnat}
\bibliography{bibliography}
\end{document}